%% file: main_V1.tex
\documentclass[10pt,twocolumn,letterpaper]{article}

\usepackage{cvpr}
\usepackage{times}
\usepackage{epsfig}
\usepackage{graphicx}
\usepackage{amsmath}
\usepackage{amssymb}

\usepackage[usenames,dvipsnames,svgnames,table]{xcolor}
\usepackage{graphicx}
\usepackage{amsmath,amssymb}
\usepackage{amsthm}
\usepackage{bm}
\usepackage{tabularx}
\usepackage{multirow}
\usepackage{hhline}
\usepackage{subfig}
\usepackage{booktabs}
\usepackage{sidecap}
\usepackage{epstopdf}
\usepackage{lipsum}

\usepackage[normalem]{ulem}

\usepackage[pagebackref=true,breaklinks=false,colorlinks,bookmarks=false]{hyperref}

\def\ie{\emph{i.e.}\xspace}
\DeclareMathOperator*{\argmax}{arg\; max}

\cvprfinalcopy 

\def\cvprPaperID{1486} 

\ifcvprfinal\fi

\begin{document}

\title{Siamese Instance Search for Tracking}

\author{Ran~Tao, Efstratios~Gavves, Arnold~W.M.~Smeulders\\
QUVA Lab,
Science Park 904, 1098 XH Amsterdam\\
{\tt\small \{r.tao,egavves,a.w.m.smeulders\}@uva.nl}
}

\maketitle
\thispagestyle{empty}

\begin{abstract}
\input{abstract_V1}
\end{abstract}

\section{Introduction}

\input{introduction_V1}

\section{Related Work}
\input{related_work_V1}
\section{Siamese Instance Search Tracker}\label{sec:model}
\input{model_V1}

\section{Experimenents}
\input{experiments_V1}

\section{Conclusion}
\input{conclusions_V1}
{\small
\bibliographystyle{ieee}
\bibliography{main_V1}
}

\end{document}

%% file: abstract_V1.tex



In this paper\footnote{This paper is accepted to CVPR 2016.} we present a tracker, which is radically different from state-of-the-art trackers: we apply no model updating, no occlusion detection, no combination of trackers, no geometric matching, and still deliver state-of-the-art tracking performance, as demonstrated on the popular online tracking benchmark (OTB) and six very challenging YouTube videos. The presented tracker simply matches the initial patch of the target in the first frame with candidates in a new frame and returns the most similar patch by a learned matching function. The strength of the matching function comes from being extensively trained generically, \ie, without any data of the target, using a Siamese deep neural network, which we design for tracking. Once learned, the matching function is used as is, without any adapting, to track previously unseen targets. It turns out that the learned matching function is so powerful that a simple tracker built upon it, coined Siamese INstance search Tracker, \emph{SINT}, which only uses the original observation of the target from the first frame, suffices to reach state-of-the-art performance. Further, we show the proposed tracker even allows for target re-identification after the target was absent for a complete video shot.

%% file: introduction_V1.tex
At the core of many tracking algorithms is the function by which the image of the target is matched to the incoming frames.
The matching function for tracking ideally provides good matching even if the target in the video is occluded, changes its scale, rotates in and out-of-plane or, undergoes uneven illumination, camera motion and other disturbing factors~\cite{smeulders2014visual, wu2013online}.
One way to proceed is to model each of these distortions explicitly in the matching by introducing affine transformations~\cite{Lucas1981}, probabilistic matching~\cite{Comaniciu2000}, eigen images~\cite{Ross2008}, illumination invariants~\cite{Nguyen2006}, occlusion detection~\cite{pan2007robust}. 
While one explicit matching mechanism may be well-fitted to solve one distortion, it is likely to disturb another.

\begin{figure}[t!]
    \centering 
    \includegraphics[width=1.0\linewidth]{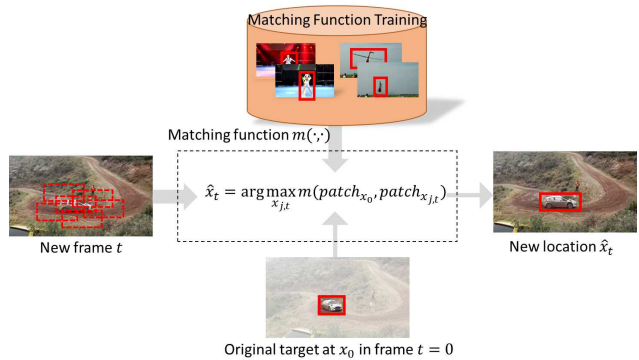}
    \caption{The tracker simply finds the patch that matches best to the original patch of the target in the first frame, using a learned matching function. The matching function is learned once on a rich video dataset. Once it has been learned, it is applied as is, without any adapting, to new videos of previously unseen target objects. We \emph{do not} apply offline target learning and the target is not included in the training video dataset.}
    \label{fig:intro}
\end{figure}

In this work, rather than explicitly modeling the matching for particular distortions, we propose \emph{to learn} the matching mechanism. More specifically, we suggest that we learn from external videos that contain various disturbing factors the invariances without, however, explicitly modeling these invariances.
If the set of external videos is sufficiently large, the goal is to learn a generically applicable matching function \emph{a priori}. We take extra care that there is absolutely no overlap between the videos we use for training and any of the tracking videos for evaluation. Namely, we \emph{do not} aim to do any offline learning of the tracking targets, since in that case we would essentially learn an object detector. Instead, in the learning we focus on the generic set of object appearance variations in videos. In this way, we optimize the matching function between an arbitrary target and patches from subsequent frames. Once the matching function has been learnt on the external data we do not adapt it anymore and, we apply it as is to new tracking videos of previously unseen target objects.

We focus on learning the matching function suited for application in trackers.
Hence, our aim is not to build a fully fledged tracker which might need explicit occlusion detection~\cite{Pernici2014}, model updating ~\cite{zhang2014meem, Holzer2015, hare2011struck}, tracker combination~\cite{zhang2014meem}, forget mechanisms~\cite{Nguyen2006, hare2011struck} and other.
We rather focus on the matching function alone, similar to the simplicity of the normalized cross-correlation (NCC) tracker~\cite{duda1973pattern, Briechle2001}. In this paper, we simply match the initial target in the first frame with the candidates in a new frame and return the most similar one by the learnt matching function, without updating the target, tracker combination, occlusion detection and alike. Figure~\ref{fig:intro} illustrates the tracking algorithm.

This approach to tracking bears some similarity to instance search~\cite{tao2014locality, tao2015attributes, philbin2007object, Tolias2015}, where the target specified in the query image is searched for in a pile of images.
Introducing matching learning~\cite{tao2015attributes} allows for accurate instance search of generic objects even when the relevant images in the search set show drastically different views of the target object from the query image.
Here we intend to learn a generic matching function to cope with all sorts of appearance variations from tracking examples.
After learning, the matching function is capable of comparing patches recorded under very different conditions for new objects, or, even for new object types that the function has not seen before.

We summarize the contributions of the work as follows. First, we propose to learn a generic matching function for tracking, from external video data, to robustly handle the common appearance variations an object can undergo in video sequences. The learnt function can be applied as is, without any adapting, to new tracking videos of previously unseen target objects. Second, on the basis of the learnt generic matching function, we present a tracker, which reaches state-of-the-art tracking performance. The presented tracker is radically different from state-of-the-art trackers. We apply no model updating, no occlusion detection, no combination of trackers, no geometric matching and alike. In each frame, the tracker simply finds the candidate patch that matches best to the initial patch of the target in the first frame by the learned matching function. Third, to learn the matching function, we use a two-stream Siamese network~\cite{bromley1993signature}, which we design specifically for tracking. Further, in the absence of any drifting that one would expect by on-the-fly model updating, the proposed tracker allows for successful target object re-identification after the target was absent for a long period of time, \eg, a complete shot.

%% file: related_work_V1.tex
\textbf{Matching functions in tracking} One of the most basic concept of tracking is the direct matching between the intensity values of the target patch and the patches taken from the incoming image. The oldest tracking algorithm does just that by normalized cross-correlation~\cite{duda1973pattern, Briechle2001}. Its simplicity is also its strength, still being in use as part of the TLD-tracker~\cite{kalal2012tracking}. Subsequent trackers have reconsidered the matching function by focusing on the various distortions to the target image faced in tracking. The Lucas and Kanade tracker~\cite{Lucas1981} adds an affine transformation to the matching function. MST~\cite{Comaniciu2000} relies on probabilistic matching. FRT~\cite{Adam2006} uses the earth mover's distance matching. And IVT~\cite{Ross2008} matches by the metric of eigen images obtained during tracking. L1T~\cite{Mei2009} is successful with L1-metric matching on graphs of fragments. SPT~\cite{Wang2011} uses super-pixels for matching, HBT~\cite{Godec2011} uses HOG-features in a probabilistic approach, and FBT~\cite{Nguyen2006} uses color invariants for robustness against illumination variations. Different from all methods above, which pursue explicit modeling of the matching function, this paper aims to \emph{learn} the matching function from example videos annotated with the correct boxes.

\textbf{Recent tracking methods} In recognition of the hardness of the task, composite trackers have been introduced. TLD~\cite{kalal2012tracking} integrates the NCC matching for recovery with a differential tracker and a complex updating model. Struck~\cite{hare2011struck} is based on structural SVM with the displacement as the continuous output, with a cautious update mechanism. More recently, MEEM~\cite{zhang2014meem} successfully learns and updates a discriminative tracker, keeping a set of historical snapshots as experts who derive the per frame prediction based on an entropy regularized optimization. Alien~\cite{Pernici2014} is a successful long-term tracker relying on oversampling of local features and RANSAC-based geometric matching. In the very recent MUSTer~\cite{Hong2015} one component stores short-term memories of the target for short-term tracking using integrated correlation filters, where the long-term memory is based on RANSAC matching again. Finally, the AND-OR tracker~\cite{Wu2015} proposes a discriminative learning of hierarchical, compositional and-or graphs that account for the appearance and structural variations of the object. In this paper, we focus on simple tracking inference scheme, namely finding the patch that matches best to the initial target in the first frame. The complexity, instead, is incorporated externally, where the matching function is trained to be robust against appearance variations. Hence, rather than learning on-the-fly, we learn what can be encountered \emph{in general} without requiring target-specific learning. Once learned, the matching function can be built in the successful, aforementioned composite trackers to enhance their performance.

\textbf{Deep learning in tracking}~\cite{wang2013learning} uses a stacked denoising autoencoder to learn tracking features. The features are performing poorly, however.~\cite{li2014deeptrack} learns a target classifier online, which is fundamentally hampered by a lack of data.~\cite{hong2015online} focuses on learning target-specific saliency map using pre-trained ImageNet network.~\cite{wang2015transferring} pre-trains a convolutional neural network for measuring generic objectness on ImageNet 2014 detection set and adapts the network online to predict the target-specific objectness. Compared to previous works, this work focuses on a different part of a tracker. We employ deep neural networks to learn a generic matching function from rich external data to compare patches for use in tracking. 

\textbf{Instance Search} Instance search from one example, also known as particular object retrieval, is related to object tracking, especially when localized~\cite{jiang2012randomized,tao2014locality}. The most popular paradigm is based on matching local image descriptors between the query and the candidate image~\cite{sivic2003video,philbin2007object,jegou2008hamming,qin2013query,tolias2013aggregate,tao2014locality} and is especially accurate for buildings~\cite{philbin2007object}. Recently, ~\cite{tao2015attributes} proposed to learn a robust representation for instance search of less textured, more generic objects, showing good accuracy despite the significant appearance changes between the query and the database images. We derive some inspiration from~\cite{tao2015attributes}. We propose to learn a robust matching function for matching arbitrary, generic objects that may undergo all sorts of appearance variations. We focus, however, on tracking. Instead of focusing on a specific category \eg, shoes, and learning from images with a white background~\cite{tao2015attributes}, we learn in this work a universal matching model suited for tracking that applies to all categories and all realistic imaging conditions.

\textbf{Siamese architecture}~\cite{bromley1993signature} proposes the two-stream Siamese architecture for signature verification. Later, the two-stream network architecture has been applied to face verification~\cite{chopra2005learning,taigman2014deepface}, ground-to-aerial image matching~\cite{lin2015learning}, local patch descriptor learning~\cite{han2015matchnet,zagoruyko2015learning} and stereo matching~\cite{vzbontar2015computing}. In this work, we design a Siamese network-architecture to learn robust and generic representation for object tracking, aiming to be invariant to all sorts of appearance variations in practical tracking scenarios.

\textbf{Fast localization} Tracking also bears resemblance to the object localization problem. Usually, it requires efficient processing of multiple regions in one frame.~\cite{li2013codemaps} proposes efficient region computation by reordering the encoding, pooling and classification steps for the `shallow' representations such as Fisher vector~\cite{perronnin2010improving}. Recent work by Girshick~\cite{girshick2015fast} proposes an efficient way of processing multiple regions in one single pass through the deep neural network for fast object detection. Inspired by~\cite{girshick2015fast}, we incorporate the region-of-interest pooling layer into our network for fast processing of multiple regions in one frame for tracking.

%% file: model_V1.tex
In the following we describe the proposed method for tracking, which is coined Siamese INstance search Tracker, SINT for abbreviation. We first present the matching function, which is the core of the tracker. Then we describe the simple online tracking inference.

\subsection{Matching Function}

To learn a matching function robust to all sorts of distortions as described earlier, we need a model that operates on pairs of data, $(x_j, x_k)$. A network architecture that has been successfully shown to work well on pairs of data is the two-stream Siamese architecture~\cite{bromley1993signature,chopra2005learning}. A Siamese architecture builds on top of convolutional networks. 
Next, we analyze the different components of the proposed two-stream network which we coin \emph{Siamese Invariance Network}.

\begin{figure}
    \centering {
        \includegraphics[width=0.9\linewidth]{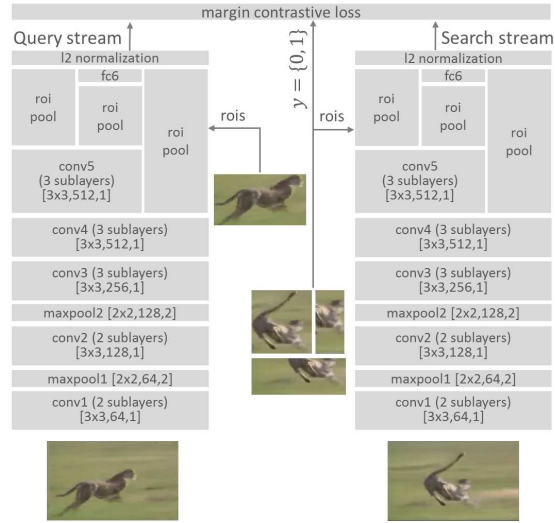}
    }
    \caption{The proposed Siamese invariance network to learn the generic matching function for tracking. `conv', `maxpool', `roipool' and `fc' stand for convolution, max pooling, region-of-interest pooling and fully connected layers respectively. Numbers in square brackets are kernel size, number of outputs and stride. The fully connected layer has 4096 units. All conv layers are followed by rectified linear units (ReLU)~\cite{nair2010rectified}.} 
    \label{fig:siamese_net}
\end{figure}

\noindent\textbf{Network architecture} We use a Siamese architecture composed of two branches. The Siamese network processes the two inputs separately, through individual networks that usually take the form of a convolutional neural network. For individual branches, we design and compare two different network architectures, a small one similar to AlexNet~\cite{krizhevsky2012imagenet} and a very deep one inspired by VGGNet~\cite{simonyan2015very} (Figure~\ref{fig:siamese_net})\footnote{Due to the space limit, only the very deep network is shown here. We put the illustration of the other AlexNet-like network in the supplementary material.}. In the following we highlight the distinctive designs of the networks as compared to the successful AlexNet and VGGNet.

Being largely a localization task the tracking problem is naturally susceptive to rough  discretizations. 
Aiming for precise localization, we design our network with very few maxing pooling layers, fewer than the networks in~\cite{krizhevsky2012imagenet,simonyan2015very}. Indeed, as max pooling maintains only the strongest of the activations from a local neighborhood to use as input for the subsequent layers, the spatial resolution of the activations is aggressively reduced, at the very least by 50\% only in the simple case of $2\times2$ local neighborhoods. An advantage of max pooling is it introduces invariance to local deformations. However, this is more important for object categorization, where the objects vary a lot in appearance. In tracking even if the target object changes its appearance over time, it still remains the same object in all frames. Moreover, it is important to be able to follow the small appearance changes, such as local deformations, of the object over time. Regarding the two architectures we propose, for the AlexNet-like small net we do not include any max pooling layer, while for the VGG-like large net, we only have two max pooling at the very early stage (see Figure~\ref{fig:siamese_net}), as the lower level layers learn filters of very small receptive fields and their max pooling layers are important to maintain robustness to local noise.

In tracking one typically needs to evaluate hundreds of candidate regions for the next frame. Although one can simply pass through the candidate regions independently, this would lead to a severe computation overhead, especially since there is a significant overlap between the candidate regions. Therefore, we employ a region pooling layer~\cite{girshick2015fast} for the fast processing of multiple overlapping regions. Each branch of the Siamese architecture takes as input one image and a set of bounding box regions. The network first processes the entire image for a few layers, then the region pooling layer converts the feature map from a particular region into a fixed-length representation. Having a fixed length representation, one can now proceed to the subsequent layers.


The layers in a deep network capture progressively more abstract representations~\cite{zeiler2014visualizing}. Typically, the filters of the lower layers get activated the most on lower level visual patterns, such as edges and angles, whereas higher layers get activated the most on more complex patterns, such as faces and wheels. Also, the deeper one layer is, the more invariant it is to appearance changes but also less discriminative, especially for instance-level distinction. In tracking we do not know the type of target object we want to track, whether it is highly textured with rich low level patterns or not. We do not know either the complexity of the background, whether there are confusing objects in which case higher discrimination would probably be more helpful. For this reason we propose to use the outputs from multiple layers as the intermediate representation that is then fed to the loss function. Similar observations have also been made in~\cite{long2015fully, hariharan2015hypercolumns} for different tasks, semantic segmentation and fine-grained localization specifically. All activations are pooled using the region pooling layers.

Given that modern convolutional neural networks use rectified linear units that do not bound the output values, the nonlinear activations can vary a lot in the range of values they produce. As such and without considerations, the network output and the loss function will be heavily influenced by the scale of the generated features and not their representation quality. To avoid this we propose to add an $\ell_2$ normalization layer before the loss layer. The normalization layer is applied on each of the layer activations that are fed to the loss layer and has the property of maintaining the direction of the feature, while forcing features from different scales to lie on the same unit sphere.

Compared to standard convolutional neural networks, AlexNet and VGGNet~\cite{krizhevsky2012imagenet, simonyan2015very}, our architecture has several differences, highlighted above. However, we also explicitly design our networks to be compatible to AlexNet and VGGNet. In this way, we are able to initialize the weights of our networks using the ImageNet-pretrained AlexNet and VGGNet to avoid training from scratch, something that would likely lead to overfitting. Last, note that we keep the parameters of the two convolutional network branches tied together, as there would be an increased danger of overfitting otherwise.\\

\noindent\textbf{Network input} Our training data consist of videos of objects, whose bounding box location is provided to us. To emulate the instance search paradigm and to avoid confusion, we coin the first stream of our network as query stream, whereas the second stream of our network as search stream. For the query stream we randomly pick one frame from the video and use the annotated patch of the target. Since we want to be robust to as many types of variations that we might face when tracking novel objects as possible, for the search stream we randomly pick another video frame that does not need to be adjacent to the frame of the query stream. From the frame of the search stream we sample boxes and the ones that overlap more than $\rho_{+}$ with the ground truth are deemed positives, while the ones that overlap less than $\rho_{-}$ with the ground truth are deemed negatives. From these we form positive and negative pairs of data that we use for the training.\\

\noindent\textbf{Loss} In the end, the two branches in the Siamese Invariance Network are connected with a single loss layer. For tracking we want the network to generate feature representations, that are close by enough for positive pairs, whereas they are far away at least by a minimum for negative pairs. Bearing these requirements in mind and inspired by~\cite{chopra2005learning}, we employ the margin contrastive loss
\begin{equation}
\begin{aligned}
\mathcal{L}(x_{j},x_{k}, y_{jk})=\frac{1}{2}y_{jk} D^{2}+\frac{1}{2}(1-y_{jk})\max(0,\epsilon-D^{2}),
\end{aligned}
\label{eqn:eqn_contrastive_loss}
\end{equation}
where $D=\|f(x_{j})-f(x_{k})\|_2$ is the Euclidean distance of two $\ell_2$-normalized latent representations, $y_{jk}\in\{0,1\}$ indicates whether $x_{j}$ and $x_{k}$ are the same object or not, and $\epsilon$ is the minimum distance margin that pairs depicting different objects should satisfy.\\

\noindent\textbf{Data} As tracking is an inherently online task, where no training data related to the target object are available, it is important to emphasize that \emph{the network is learnt on external videos that do not appear in the tracking evaluation sets}. The data should be varying enough, covering a good amount of semantics and not focus on particular objects, otherwise the tuned network parameters will overfit to particular object categories. Furthermore, as we do not explicitly learn types of invariances, namely we do not learn ``illumination invariance'' separately from ``scale invariance'', therefore in the external data we do not need any specific variation labels. The only requirement is the box annotations within the video following a particular object. 


\subsection{Tracking Inference}

Once we have completed the learning of the matching function, we are ready to deploy it as is to tracking, without any further adapting.
We propose a simple tracking strategy.
As the only reliable data we have for the target object is its location at the first frame, at each frame we compare the sampled candidate boxes with the target object at the first frame.
We pass all the candidate boxes from the search stream of our network and pick the candidate box that matches best to the original target,
\begin{equation}
\hat{x}_{t} = \argmax_{x_{j, t}} m(x_{t=0}, x_{j, t}),
\end{equation}
where $x_{j, t}$ are all the candidate boxes at frame $t$, $m$ is the learned matching function, $m(x,y)=f(x)^{T}f(y)$.\\

\noindent\textbf{Candidate sampling}
We employ the radius sampling strategy~\cite{hare2011struck}.
More specifically, around the predicted location of the previous frame we sample locations evenly on  circles of different radii. Different from~\cite{hare2011struck}, to handle scale variations we generate at each sampled location multiple candidate boxes at different scales.

\noindent\textbf{Box refinement}
Provided that the box prediction is accurate enough, ~\cite{felzenszwalb2010object,girshick2014rich} showed that a refinement step of the boxes can improve localization accuracy significantly.
To this end we adopt their strategy and refine at each frame the predicted bounding box further. 

As in~\cite{girshick2014rich} we train four Ridge regressors for the $(x,y)$ coordinates of the box center and the width and height $(w,h)$ of the box based on the first frame. The regressors are not updated during tracking in order to avoid the risk of contaminating the regressors with noisy data. For each frame, the regressors take the representation of the picked candidate box as input and produce a refined box.

%% file: experiments_V1.tex
\subsection{Implementation Details}
\textbf{Candidate Sampling} We use the radius sampling strategy~\cite{hare2011struck} to generate candidate boxes. We use 10 radial and 10 angular divisions. The search radius is set to be the longer axis of the initial box in the first frame. At each sample location, we generate three scaled versions of the initial box with the scales being $\{\frac{\sqrt2}{2}, 1, \sqrt2\}$. 

\textbf{Network training} We use the ALOV dataset~\cite{smeulders2014visual} for training and validation. We choose ALOV for training as it covers many types of variations one could expect in tracking. \emph{We exclude the 12 videos in ALOV that are also in tracking benchmark (OTB)~\cite{wu2013online}, as we evaluate the proposed tracker on OTB.} After removing the 12 videos, the training set and the tracking evaluation set have no common objects. 
From every two frames in a video, we generate multiple pairs. One element in a pair is the groundtruth bounding box in one frame and the other element is a box sampled in the other frame. The pair is considered to be positive if the sampled box has a intersection-over-union overlap larger than 0.7 with the corresponding groundtruth box and considered to be negative if the overlap is smaller than 0.5. 
The training pairs and validation pairs are generated from different videos, and therefore from different objects. For training, in total we have sampled from ALOV dataset $60,000$ pairs of frames and each pair of frames has 128 pairs of boxes. For validation, we have gathered $2,000$ pairs of frames and the same as for training each pair of frames contains 128 pairs of boxes.

Instead of training the two-stream Siamese network from scratch, we load the pre-trained network parameters and fine tune the Siamese network. Specifically, we use the networks pre-trained for ImageNet classification, available in the Caffe library~\cite{jia2014caffe}. The initial fine tuning learning rate is $0.001$ and the weight decay parameter is $0.001$. The learning rate is decreased by a factor of 10 after every 2 epochs. We stop tuning when the validation loss does not decrease any more.


\subsection{Dataset and evaluation metrics}

\noindent\textbf{Dataset} To evaluate the tracking performance, we use the online tracking benchmark (OTB)~\cite{wu2013online}. OTB is a collection of 50 videos. 51 tracking sequences are defined with bounding box annotations. The dataset covers various challenging aspects in object tracking, such as fast motion, deformation, background clutter and occlusion.      

\noindent\textbf{Evaluation metrics} We follow the evaluation protocol of~\cite{wu2013online}, where two metrics are used: success plot and precision plot. Both metrics measure the percentage of successfully tracked frames. For the success plot, a frame is declared to be successfully tracked if the estimated bounding box and the groundtruth box have an intersection-over-union overlap larger than a certain threshold. For precision plot, tracking on a frame is considered successful if the distance between the centers of the predicted box and the groundtruth box is under some threshold. A plot is given by varying the threshold values. Tracking algorithms are ranked based on the area under curve (AUC) score for the success plot and precision at threshold 20 (Prec@20) for the precision plot. We use the available toolkit provided by the benchmark to generate plots and numbers. In the following, we also use success rate where needed, \ie, the percentage of successfully tracked frames.

\subsection{Design evaluation}


We first validate our design choices of the network. In this sets of experiments, box refinement is not considered.\\

\begin{table}
    \renewcommand{\arraystretch}{1.0}
    \centering
    \scalebox{0.75} {
    \setlength{\tabcolsep}{6pt}
    \begin{tabular}{lcc}
        \toprule
            & AUC & Prec@20  \\
        \midrule
           \emph{(a)} pretrained-alexnet-fc6 & 42.8 & 66.3 \\
           \emph{(b)} firstframe-Siamese-finetuned-alexnet-fc6 & 44.0 & 67.9 \\
           \emph{(c)} Siamese-finetuned-alexnet-fc6 & 47.4 & 72.0 \\
           \emph{(d)} pretrained-alexnet-fc6-nomaxpooling & 50.0 & 70.8 \\
           \emph{(e)} Siamese-finetuned-alexnet-fc6-nomaxpooling & 53.9  & 74.8 \\
           \emph{(f)} Siamese-finetuned-alexnet-conv45fc6-nomaxpooling & 55.0 & 76.2 \\
           \emph{(g)} Siamese-finetuned-vgg16-conv45fc6-nomaxpooling & 59.2 & 83.6 \\
        \bottomrule
    \end{tabular}
    }
    \caption{Evaluation of different architectural and design choices of the Siamese invariance network for tracking on the OTB dataset~\cite{wu2013online}. We use the recommended evaluation methods, namely the area under the curve (AUC) for the success plot and the precision at 20 (Prec@20) for the precision plot.}
    \label{tab:tab_design}
    \vspace{-2mm}
\end{table}

\noindent\textbf{Network tuned generically on external video data vs. network pre-tuned on ImageNet vs. network fine tuned target-specifically on first frame}
In this experiment, we show the effectiveness of the Siamese network tuned on external data. To that end, we compare the Siamese fine tuned AlexNet-style network using ALOV (denoted as ``Siamese-finetuned-alexnet-fc6'') with the ImageNet pre-tuned AlexNet (``pretrained-alexnet-fc6'') and the Siamese fine tuned network using the training pairs gathered in the first frame (``firstframe-Siamese-finetuned-alexnet-fc6''). In this comparison, all three use a single layer \textit{fc6} for feature representation. As shown in the rows \emph{(a)}-\emph{(c)} of Table~\ref{tab:tab_design}, the Siamese fine tuned network using ALOV~\emph{(c)} significantly improves over the pre-tuned net~\emph{(a)}, while fine tuning on the first frame~\emph{(b)} gives a marginal improvement. We conclude that Siamese networks fine tuned using large amount of external data are to be preferred.\\

\begin{figure*}[t!]
    \centering {
        \includegraphics[width=0.85\linewidth]{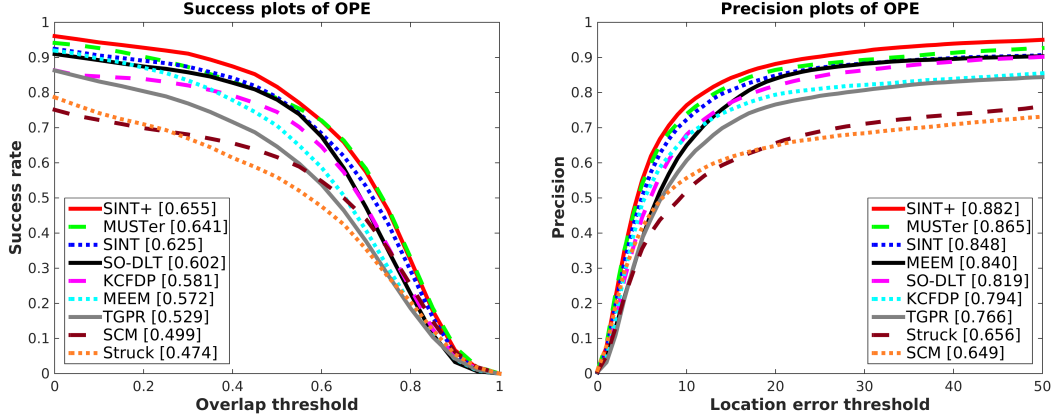}
    }
    \caption{State-of-the-art comparison on OTB~\cite{wu2013online}. In spite of the fact that the online part of the proposed SINT is just selecting the patch that matches best to the target in the first frame, SINT is on par with state-of-the-art tracker. SINT+, using a better candidate sampling than SINT and optical flow as an additional component, achieves the best performance.}
    \label{fig:benchmark_eval}
\end{figure*}

\noindent\textbf{To max pool or not to max pool?}
We now examine our design choice of having no maxing pooling layers in the network (``pretrained-alexnet-fc6-nomaxpooling'' vs. ``pretrained-alexnet-fc6'' and ``Siamese-finetuned-alexnet-fc6-nomaxpooling'' vs. ``Siamese-finetuned-alexnet-fc6'').
As shown in Table~\ref{tab:tab_design}, \emph{(d)} vs. \emph{(a)} and \emph{(e)} vs. \emph{(c)}, including max pooling layers deteriorates accuracy, as expected due to the reduction of the resolution of the feature maps which causes poor localization.
When inspecting the results when no max pooling layers are included, the success rate improvement is higher at higher intersection-over-union overlap ratios, see Table~\ref{tab:no_maxpool}. We conclude that max pooling layers are not necessary for our Siamese invariance network with small AlexNet-style architecture.\\


\noindent\textbf{Multi-layer features vs. single-layer features}
Next, we evaluate whether it is more advantageous to use features from a single layer or from multiple layers.
We compare ``Siamese-finetuned-alexnet-conv45fc6-nomaxpooling'', which uses the outputs of layers \textit{conv4}, \textit{conv5} and \textit{fc6} as features, with ``Siamese-finetuned-alexnet-fc6-nomaxpooling'', which uses the output of \textit{fc6} as feature.
Table~\ref{tab:tab_design} shows that using multi-layer features is helpful (\emph(e) vs. \emph(f)). 
We conclude that using features from multiple layers is advantageous.\\

\noindent\textbf{Large net vs. small net} Lastly, we compare a VGGNet-style architecture with an AlexNet-style architecture (``Siamese-finetuned-vgg16-conv45fc6-nomaxpooling'' vs. ``Siamese-finetuned-alexnet-fc6-nomaxpooling'').
Both use as features the outputs of three layers.
As shown in the last two rows \emph{(f)} and \emph{(g)} of Table~\ref{tab:tab_design}, using a deeper network improves the performance significantly.

\begin{table}
    \renewcommand{\arraystretch}{1.0}
    \centering
    \scalebox{0.72} {
    \setlength{\tabcolsep}{6pt}
    \begin{tabular}{lccc}
        \toprule
            & sr@0.3 & sr@0.5 & sr@0.7 \\
        \midrule
        	pretrained-alexnet-fc6 & 68.3 & 46.2 & 19.6 \\
        	pretrained-alexnet-fc6-nomaxpooling & 75.3 & 58.1 & 32.6 \\ 

        \midrule
        	Siamese-finetuned-alexnet-fc6 & 74.6 & 56.2 & 25.4 \\
        	Siamese-finetuned-alexnet-fc6-nomaxpooling & 79.3  & 67.6  & 38.8  \\
        \bottomrule
    \end{tabular}
    }
    \caption{Success rates (sr) of the tracker at three intersection-over-union overlap ratios for different network architectures. From the table it is clear that a network architecture without max pooling delivers a more precise localization and hence a better matching function.}
    \label{tab:no_maxpool}
    \vspace{-2mm}
\end{table} 



\subsection{State-of-the-art comparison}


\noindent\textbf{Overall comparison} In addition to the 29 trackers included in the benchmark~\cite{wu2013online}, \eg, TLD~\cite{kalal2012tracking}, Struck~\cite{hare2011struck} and SCM~\cite{zhong2012robust}, we also include the most recent trackers for comparison. The included recent trackers are TGPR~\cite{gao2014transfer}, MEEM~\cite{zhang2014meem}, SO-DLT~\cite{wang2015transferring}, KCFDP~\cite{Huang2015Enable} and MUSTer~\cite{Hong2015}.

As described earlier, the proposed SINT focuses on the tracking matching function, while having a simple online inference. As a preliminary demonstration that SINT can be further improved by employing more advanced online components, we also evaluate a variant of SINT, coined SINT+, which uses an adaptive candidate sampling strategy suggested by~\cite{wang2015understanding} and optical flow~\cite{brox2011large}. In SINT+, the sampling range is adaptive to the image resolution, set to be $30/512*w$ in this experiment, where $w$ is the image width. Optical flow is used in SINT+ to filter out motion inconsistent candidates. Specifically, given the pixels covered by the predicted box in the previous frame and the estimated optical flow, we know where those pixels are in the current frame and we remove the candidate boxes that contain less than 25\% of those pixels, as these candidates are deemed inconsistent to the motion.

Figure~\ref{fig:benchmark_eval} shows the overall performance. For clarity, only the top performing trackers are shown.
Despite relying on a simple NCC-like tracking inference, SINT reaches state-of-the-art performance, being tantalizingly close to MUSTer~\cite{Hong2015} and more accurate than others by a considerable margin. SINT+, with an adaptive sampling and a simple use of optical flow, further improves SINT, outperforming clearly all state-of-the-art other trackers.\\ 

\noindent\textbf{Temporal and spatial robustness} To verify the robustness of the proposed tracker, we conduct the temporal robustness evaluation (TRE) and spatial robustness evaluation (SRE) defined by the benchmark. The results are summarized in Table~\ref{tab:TRE_SRE}. Compared to MEEM and MUSTer, SINT is temporally and spatially the same as robust, if not better.\\

\begin{table}
    \renewcommand{\arraystretch}{1.0}
    \centering
    \scalebox{0.9} {
    \setlength{\tabcolsep}{6pt}
    \begin{tabular}{lccc}
        \toprule
            & OPE & TRE & SRE \\
        \midrule
        MEEM &57.2 / 84.0 & 58.5 / 83.2 & 51.8 / 76.9\\
        MUSTer & 62.1 / 83.6 & 60.9 / 81.1 & 56.2 / 78.9 \\
        SINT & 62.5 / 84.8 & 64.3 / 84.9 &  57.9 / 80.6 \\
        \bottomrule
    \end{tabular}
    }
    \caption{Robustness evaluation on OTB, measured in AUC/Prec@20. OPE is one-pass evaluation. TRE and SRE are temporal and spatial robustness evaluation. The results of MEEM are taken from~\cite{zhang2014meem} and the results of MUSTer are obtained using the publicly available code.}
    \label{tab:TRE_SRE}
    \vspace{-3mm}
\end{table}

\begin{figure}

    \centering {

        \includegraphics[width=0.78\linewidth]{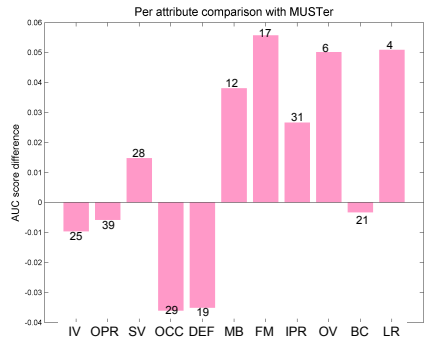}
    }
    \vspace{-3mm}
    \caption{Per attribute comparison on AUC score of the proposed SINT with MUSTer~\cite{Hong2015}. The bars are the performance difference between SINT and MUSTer. Positive means SINT is better. The integer number at each bar is the number of tracking sequences belonging to that group.}
    \label{fig:per_att}
     \vspace{-2mm}
\end{figure}

\noindent\textbf{Per distortion type comparison} Further, the 50 videos in the benchmark are annotated with 11 distortion types(\eg, illumination variation, occlusion \etc).
To gain more insights, we evaluate the performance of SINT for individual attributes and compare with MUSTer~\cite{Hong2015}. 
SINT performs better in 6 and 7 out of the 11 groups for the AUC and the Prec@20 metrics respectively. Due to the space limit, Figure~\ref{fig:per_att} only shows the plot for AUC. It is observed that MUSTer is better mainly in ``occlusion'' and ``deformation'', whereas SINT is better in ``motion blur'', ``fast motion'', ``in-plane rotation'', ``out of view'' and ``low resolution''.\\

\begin{figure}

    \centering {

        \includegraphics[width=1.0\linewidth]{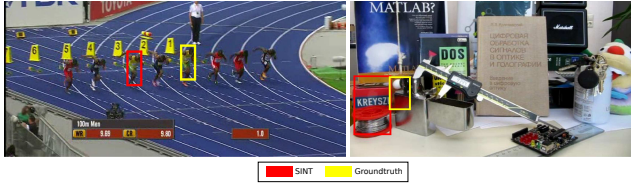}
    }
    \vspace{-4mm}
    \caption{Failure cases of SINT: similar confusing object (left) and large occlusion (right). Examples are from OTB sequences `Bolt' and `Lemming' respectively. In the left example, the tracker fires on another Jamaican runner in the same uniform as the target. In the right example, the target is heavily occluded by the lighter.}
    \label{fig:failure_cases}
\end{figure}

\noindent\textbf{Failure modes of SINT} When similar objects appear in view, the tracker may jump from the target to another as it only looks for the maximum similarity with the original patch of the target in the first frame (Figure~\ref{fig:failure_cases}: left). And, when there is large occlusion, the matching function might suffer (Figure~\ref{fig:failure_cases}: right). 

\subsection{Additional sequences and re-identification}\label{sec:add_seqs}

\begin{figure}
    \centering {
        \includegraphics[width=1\linewidth]{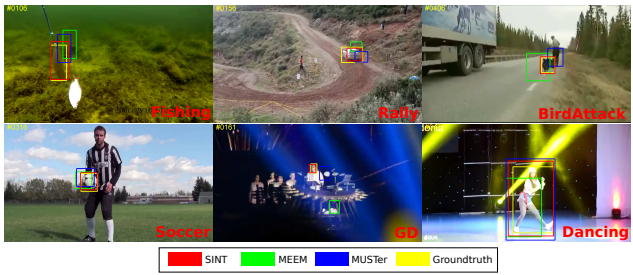}
    }
    \vspace{-4mm}
    \caption{Example frames from the 6 test sequences.}
    \label{fig:add_seqs}
\end{figure}

We now further illustrate the strength of the proposed SINT on 6 newly collected sequences from YouTube. We downloaded the sequences so that they are extra challenging in terms of tracking distortions as defined by~\cite{wu2013online}. 
Figure~\ref{fig:add_seqs} shows example frames from these sequences. The sequences have considerable degrees of scale change (``Fishing'', ``Rally'', ``BirdAttack'' and ``GD''), fast motion (``BirdAttack'', ``Soccer'' and ``Dancing''), out-of-plane rotation (``Rally'' and ``Dancing''), non-rigid deformation (``Fishing'', ``BirdAttack'' and ``Dancing''), low contrast (``Fishing''), illumination variation (``GD'' and ``Dancing'') and poorly textured objects (``Fishing'' and ``BirdAttack'').

We evaluate the proposed tracker, SINT, with MEEM~\cite{zhang2014meem} and MUSTer~\cite{Hong2015} on these sequences. The performance is summarized in Table~\ref{tab:additional_seqs}, where we adopt the AUC score metric from the benchmark~\cite{wu2013online}.
Results show that SINT is again a competitive tracker, outperforming MUSTer~\cite{Hong2015} and MEEM~\cite{zhang2014meem}.

\begin{table}
    \renewcommand{\arraystretch}{1.0}
    \centering
    \scalebox{0.7} {
    \setlength{\tabcolsep}{6pt}
    \begin{tabular}{lccc}
        \toprule
            & MEEM~\cite{zhang2014meem} & MUSTer~\cite{Hong2015} & SINT \\
        \midrule
        \emph{Fishing} & 4.3 & 11.2 & 53.7\\
        \emph{Rally} & 20.4 &  27.5 & 53.4\\
        \emph{BirdAttack}&  40.7 & 50.2 & 66.7\\
        \emph{Soccer} & 36.9 &  48.0 & 72.5 \\
        \emph{GD} & 13.8 & 34.9 & 35.8 \\
        \emph{Dancing} & 60.3 & 54.7 & 66.8 \\
        \midrule
        mean & 29.4 & 37.8 & 58.1\\
        \bottomrule
    \end{tabular}
    }
    \caption{Comparison on AUC score of the proposed SINT with MEEM~\cite{zhang2014meem} and MUSTer~\cite{Hong2015}.}
    \label{tab:additional_seqs}
\end{table} 



\begin{figure}
    \centering {
        \includegraphics[width=1\linewidth]{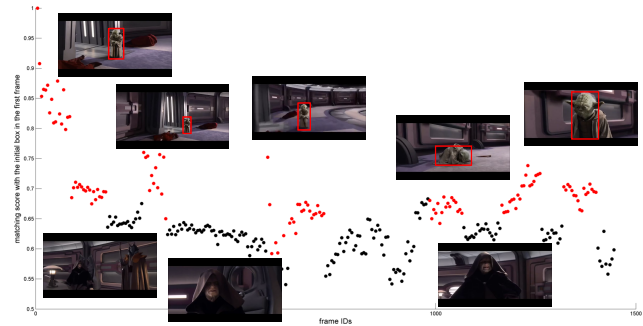}
    }
    \vspace{-3mm}
    \caption{The capability of the tracker to re-discover the target, illustrated on a 1500-frame, 12-shot \emph{Star Wars} video. One object (\emph{Yoda}) is appearing in 6 of the shots, while being absent in the intermediate ones. Red dots indicate \emph{Yoda} is present while black dots indicate \emph{Yoda} is absent. Y-axis is the matching score with the target at the first frame. The results show good capability of the tracker to discover the target when it re-enters the scene.}
    \label{fig:Yoda}
    \vspace{-3mm}
\end{figure}

We, furthermore, observe that provided a window sampling over the whole image using~\cite{zitnick2014edge}, SINT is accurate in target re-identification, after the target was missing for a significant amount of time from the video.
We illustrate this in Figure~\ref{fig:Yoda}, where we track \emph{Yoda}. As shown in Figure~\ref{fig:Yoda}, the tracker has good capability of discovering the target when it re-enters the scene after being absent for a complete shot. 

%% file: conclusions_V1.tex



This work presents Siamese INstance search Tracker, SINT. It tracks the target, simply by matching the initial target in the first frame with candidates in a new frame and returns the most similar one by a learned matching function. The strength of the tracker comes from the powerful matching function, which is the focus of the work. The matching function is learned on ALOV~\cite{smeulders2014visual}, based on the proposed two-stream very deep neural network. We take extra care that there is absolutely no overlap between the training videos and any of the videos for evaluation. Namely, we \emph{do not} aim to do any pre-learning of the tracking targets. Once learned, the matching function is used as is, without any adapting, to track arbitrary, previously unseen targets. It turns out the matching function is very effective in coping with common appearance variations an object can have in videos. The simple tracker built upon the matching function, reaches state-of-the-art performance on OTB~\cite{wu2013online}, without updating the target, tracker combination, occlusion detection and alike. Further, SINT allows for target re-identification after the target was absent for a complete shot, demonstrated on a 1500-frame, 12-shot \emph{Star Wars} video.\\ 
\\